\documentclass[10pt,twocolumn,letterpaper]{article}

\usepackage{iccv}
\usepackage{times}
\usepackage{epsfig}
\usepackage{graphicx}
\usepackage{amsmath}
\usepackage{amssymb}
\usepackage[linesnumbered,boxed,ruled,commentsnumbered]{algorithm2e}

\usepackage[breaklinks=true,bookmarks=false]{hyperref}

\iccvfinalcopy 

\def\iccvPaperID{****} 

\begin{document}

\title{Generic Multiview Visual Tracking}

\author{
Minye Wu\\
Shanghaitech University\\
{\tt\small wumy@shanghaitech.edu.cn}
\and
Haibin Ling\\
Temple University\\
{\tt\small hbling@temple.edu}
\and
Ning Bi\\
Shanghaitech University\\
{\tt\small bining@shanghaitech.edu.cn}
\and
Shenghua Gao\\
Shanghaitech University\\
{\tt\small gaoshh@shanghaitech.edu.cn}
\and
Hao Sheng\\
Beihang University\\
{\tt\small shenghao@buaa.edu.cn}
\and
Jingyi Yu\\
Shanghaitech University\\
{\tt\small yujingyi@shanghaitech.edu.cn}
}
\maketitle

\begin{abstract}
    Recent progresses in visual tracking have greatly improved the tracking performance. However, challenges such as occlusion and view change remain obstacles in real world deployment. A natural solution to these challenges is to use multiple cameras with multiview inputs, though existing systems are mostly limited to specific targets (\eg human), static cameras, and/or camera calibration.
    To break through these limitations, we propose a \textbf{generic multiview tracking} (GMT) framework that allows camera movement, while requiring neither specific object model nor camera calibration. A key innovation in our framework is a cross-camera \textbf{trajectory prediction network} (TPN), which implicitly and dynamically encodes camera geometric relations, and hence addresses missing target issues such as occlusion. Moreover, during tracking, we assemble information across different cameras to dynamically update a novel \textbf{collaborative correlation filter} (CCF), which is shared among cameras to achieve robustness against view change. The two components are integrated into a correlation filter tracking framework, where the features are trained offline using existing single view tracking datasets. For evaluation, we first contribute a new \textbf{generic multiview tracking dataset} (GMTD) with careful annotations, and then run experiments on GMTD and the PETS2009 datasets. On both datasets, the proposed GMT algorithm shows clear advantages over state-of-the-art ones.
\end{abstract}

\section{Introduction}

\begin{figure}[t]
\begin{center}
   \vspace{-2mm}\includegraphics[width=1\linewidth]{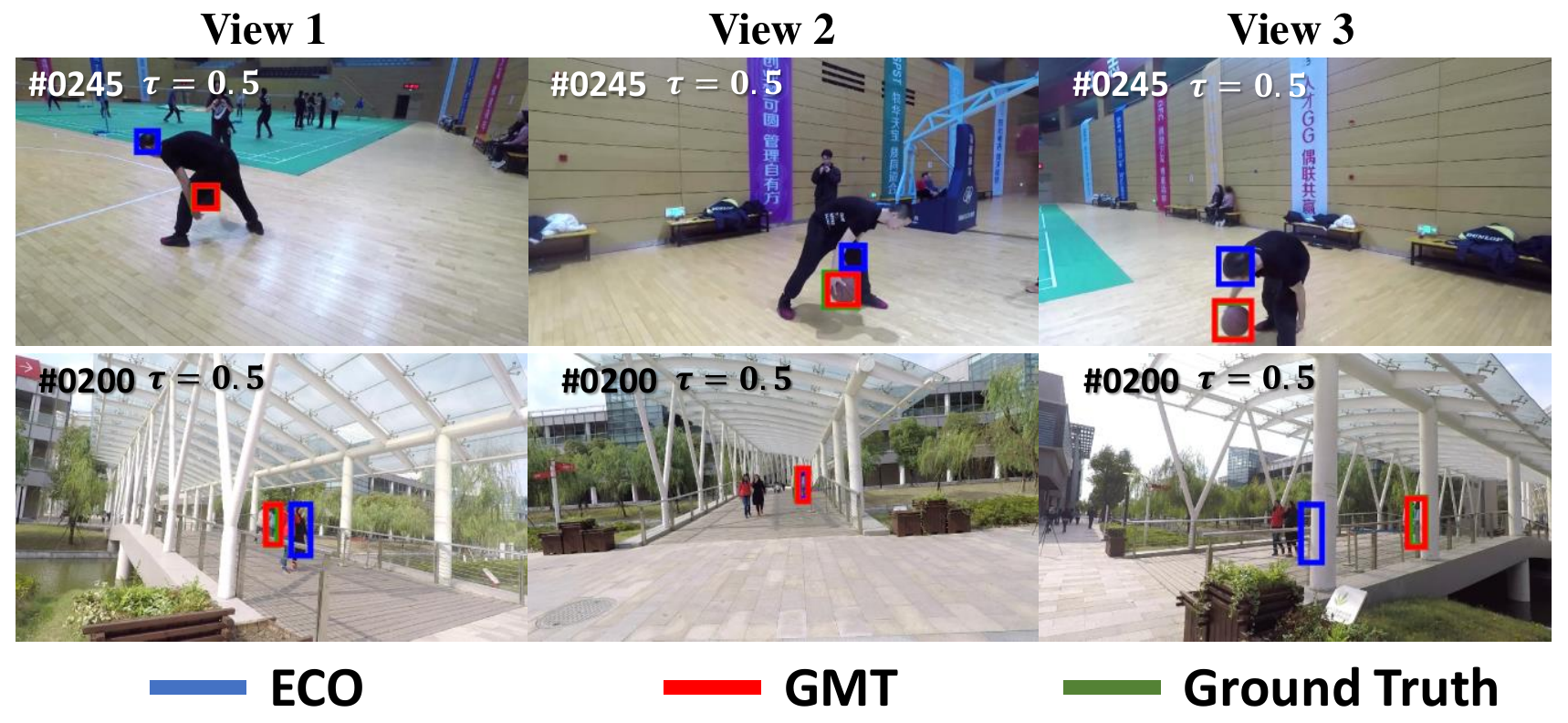}
\end{center}
   \caption{Generic multiview tracking: examples from our GMTD dataset including two frames for each of three cameras, along with tracking results by the proposed GMT algorithm and ECO.}
\label{fig:GMTD}
\vspace{-4mm}
\end{figure}

Visual object tracking is a fundamental problem in computer vision. Among different tracking tasks, we focus on generic ({\it aka} model-free) visual tracking, which requests little prior information about the tracking target and has been intensively researched due to its wide range of applications. Despite great advances in tracking algorithms, tracking in real world is still challenging especially when target appearance is distorted or damaged due to view change or occlusion. 

A natural way to alleviate the above issues is to use multiple cameras for tracking, which provides important multiview information for handling cross-view target appearance change and occlusion. Existing multiview tracking algorithms (Sec.~\ref{sub:related.mvt}), however, typically focus on specific targets like humans, and often rely heavily on detection or re-identification models. Another limitation is that cameras are often assumed to be static, so background subtraction and/or camera calibration can be used to facilitate target localization. These limitations largely restrict the generalization of multiview tracking algorithms to real world applications. Consequently, effectively using multiple cameras for generic visual tracking remains an open problem.

To address the above mentioned issues, we propose a novel {\it generic multiview tracker} (GMT) in this paper by encoding rich multiview information with learning-based strategies. A key innovation in GMT is a cross-camera {\it trajectory prediction network} (TPN), which takes tracking results from reliable views to predict those for unreliable ones. TPN effectively alleviates the problem caused by occlusion or serious target view change. Another novel component in our GMT is the {\it collaborative correlation filter} (CCF), which assembles cross-camera information to update a correlation filter shared among different views, and hence improves tracking robustness against view change. TPN and CCF are integrated into the correlation filter tracking framework, where the features are trained offline using existing single view tracking datasets for further improvement.

For evaluation, we first construct a new {\it generic multiview tracking benchmark} (GMTD) to address the scarcity of such benchmarks. Then, the proposed GMT algorithm is tested on this dataset and the PETS2009 dataset~\cite{ellis2009pets2009}, and demonstrates clear advantages in comparison with state-of-the-art tracking algorithms.

To summarize, we make the following contributions:
\begin{itemize}
    \item\vspace{-2.mm} The {\it learning-based generic multiview tracking framework}, which requires little prior information about the tracking target, allows camera movement and requires no camera calibration. 
    \item\vspace{-2.5mm} The novel {\it cross-view trajectory prediction network} that encodes camera geometric relations for improving tracking robustness. 
    \item\vspace{-2.5mm} The {\it collaborative correlation filter} that learns an online cross-view model and hence achieves natural robustness against view change.
    \item\vspace{-2.5mm} A new generic multiview tracking dataset with manual annotations per frame, which is expected to further facilitate research in related topics.
\end{itemize}
\vspace{-2.mm}The source code and the dataset will be released to public with the publication of this paper.

\section{Related Work}

\subsection{Generic Single View Visual Tracking}


Visual object tracking is one of the most important tasks in computer vision and has attracted a great amount of research efforts. It is beyond the scope of this paper to review all previous work in tracking. In the following section, we choose to review some of the most relevant ones. 

Our work is most related to correlation filter-based trackers. Based on Discriminative Correlation Filters (DCF), MOSSE is proposed in~\cite{bolme2010visual} to efficiently train the correlation filter by minimizing the sum of the squared error between ground-truth and output in Fourier domain. The idea is later adopted and extended in Kernel Correlation Filter tracking (KCF)~\cite{henriques2015high} and since then starts to gain great amount of attention. Among many tracking algorithms along the line, the series proposed by Danelljan et al.~\cite{danelljan2017discriminative,danelljan2016beyond,danelljan2017eco,bhat2018unveiling} provide the main backbone to our study. In these studies, DSST~\cite{danelljan2014accurate} and fDSST~\cite{danelljan2017discriminative} use multiple correlation filters to estimate object translation and scale separately. CCOT~\cite{danelljan2016beyond} further enhanced predictions by learning from multi-resolution feature maps. ECO~\cite{danelljan2017eco} and ECO+~\cite{bhat2018unveiling} make further improvements on feature representation, sample space management and online update scheme in order to obtain more intuitive tracker and win both accuracy and efficiency.

Another line of tracking algorithms, from which we borrow the initial feature representation, are the Siamese network-based trackers. Siamese Fully Convolutional networks (SiameseFC)~\cite{bertinetto2016fully} solves the problem by training a fully convolutional Siamese architecture to evaluate the semantic similarity between proposals and target image. Utilizing lightweight CNN network with correlation filters, DCFNet~\cite{wang2017dcfnet} performs real-time offline tracking. Recent extensions continuously improve the performance such as in~\cite{chen2018once,wang2018sint++,tao2016siamese,FanLing19cvpr}. 

Moreover, our work explicitly addresses occlusion and target view change, which have been studied explicitly as well in some single view tracking algorithms.
One way to resolve the appearance variation of the target is maintaining effective sample sets\cite{danelljan2017eco,ma2015long,ma2018adaptive}, which involves balancing different aspects of the target, for correlation filters online-training. Another strategy is conducting complementary information, for example, \cite{sun2018learning} makes use of spatial information, \cite{zhu2018end} enrolls flow between frames as part of features, \cite{he2018twofold} add a semantic branch to enhance prediction.
For occlusion, there are part-based correlation filter trackers like~\cite{chen2017visual,chen2016robust,li2015reliable}, which learn target appearance in parts and
tend to be robuster to partial-occlusion. When targets have strong structure relation, like pedestrians in RPAC~\cite{liu2015real}, KCFs are assigned to five different parts of each target for robustness. 
\cite{ma2015long, ma2018adaptive, dong2017occlusion}~set thresholds to evaluate the results of correlation filtering. 
Mei et al.~\cite{MeiLWBB11cvpr} investigate sparse representation for occlusion detection. MUSTer~\cite{guan2019real} avoid drifting and stabilize tracking by aggregating image information using short- and long-term stores. 

Different from all above-mentioned methods, our study focuses on multiview visual tracking that takes input streams from multiple cameras simultaneously. While borrowing some components from these single view tracking algorithms, we develop novel strategies such as cross-view trajectory prediction and a multiview collaborative correlation filter. These strategies, as demonstrated in our carefully designed experiments, clearly improve the tracking robustness.

\subsection{Multiview Visual Tracking}\label{sub:related.mvt}

Multi-camera inputs have been used for visual tracking. For examples, Khan et al.~\cite{Khan2006A} apply a planar homography constraint on calibrated cameras for tracking pedestrians on the ground, and show the power of multiview system with common overlaps; studies in~\cite{Kuo2010Inter,Makris2004Bridging} explore spatial relations of target object in multi-camera system by analyzing entry/exit rates across pairs of cameras; \cite{Ayazoglu2011Dynamic} exploit dynamical and geometrical constraints among static cameras in a linear model; \cite{Ristani2018Features} based on detection and re-identification methods for multi-target tracking;
the multiview trackers \cite{wen2017multi, le2018online, li2016multi, mei2015robust} track a target in each view and match each instances between cameras that are intuitively capable of occlusion situations due to diversity of observation directions. %

\begin{figure*}[t]
\begin{center}
   \includegraphics[width=1\linewidth]{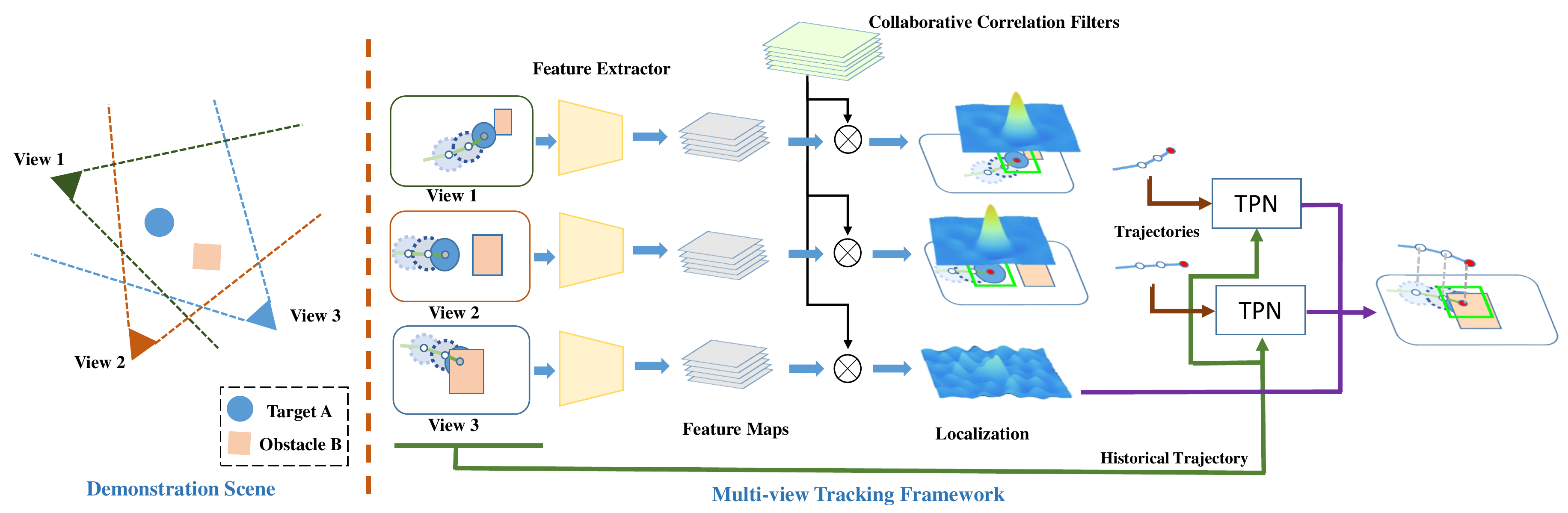}
\end{center}
   \caption{Overview. \textbf{Left}: an illustrative scene including three cameras/views (view 1, view 2 and view 3), a tracking target A and an obstacle B. \textbf{Right}: frames from all three views, where occlusion happens in view 3, serve as the input for our tracking algorithm. Three major steps are applied sequentially: \textbf{1.)} Shared feature extraction layers are performed on each view to extract cross-scale spatial-aware features (\S\ref{featureExtactSec}). \textbf{2.)} An online updated set of collaborative correlation filters are shared by all views for tracking inference (\S\ref{ccfSec}). \textbf{3.)} For a view with low tracking confidence (e.g. due to occlusion in view 3), our framework triggers trajectory prediction network (TPN) to estimate its target location based on trajectories from other views (\S\ref{TPSec}).}
\label{fig:overall}
\end{figure*}

\section{Generic Multiview Tracking Framework}

\subsection{Problem Formulation}

Given a set of synchronized video streams from different cameras/views, we aim to localize the target (initialized in the first frames) across time. More specifically, let the system input be $\mathcal{I}_t=\{I_t^c\}_{c=1}^{n_c}$ at time $t$ for $n_c$ cameras, and let $\mathcal{B}_1=\{{\bf b}_1^c \in \mathbb{R}^4\}_{c=1}^{n_c}$ be the initial target bounding boxes for all views. Then, our multiview tracking task is to localize the target by finding $\mathcal{B}_t=\{{\bf b}_t^c \in \mathbb{R}^4\}_{c=1}^{n_c}$, given $\{\mathcal{I}_1,\mathcal{I}_2,\dots,\mathcal{I}_t\}$ and $\mathcal{B}_1$,\footnote{The estimated $\{\mathcal{B}_2,\dots,\mathcal{B}_{t-1}\}$ can be used as well.} where ${\bf b}_t^c$ is the target bounding box with four parameters in view $c$ at time $t$.  
We also define a trajectory set $\mathcal{G}_{t_1,t_2}=\{\mathcal{G}_{t_1,t_2}^c\}_{c=1}^{n_c}$, where $\mathcal{G}_{t_1,t_2}^c = \{{\bf g}_t^c\in\mathbb{R}^2\}_{t=t_1}^{t_2} $ for each view $c$ and ${\bf g}_t^c$ is the center of ${\bf b}_t^c$ in a consecutive time period from time $t_1$ to time $t_2$.


\subsection{Framework Overview}

The key motivation of our {\it generic multiview tracker} (GMT) is to explore rich cross-view information to improve tracking robustness, especially against occlusion and target/camera view change. We adopt the correlation filter-based tracking framework as the backbone, and equip it with the novel {\it collaborative correlation filter} (CCF) and cross-view {\it trajectory prediction network} (TPN) techniques. An overview of pipeline of GMT is given in Figure~\ref{fig:overall}, and we briefly describe it as follows.

{During online tracking}, for newly arrived multiview images $\mathcal{I}_t$, GMT locates the target ({\it i.e.}, calculates $\mathcal{B}_t$) in three major steps. {\bf First}, for each view $c$ and scale $k$, a region of interest (RoI) patch $U_k^c$ is prepared around ${\bf b}_{t-1}^c$. The patch is then fed into the feature extraction network $\phi(\cdot)$, which is shared among different views, to generate feature maps, denoted by $X_k^c$, for each view $c$.

{\bf Second}, each feature map is convolved with the shared CCF $f$ for initial target localization, producing confidence map $Y_k^c$. The maximum response over different scales $k$ is then picked for the initial tracking results. For sufficiently confident results, they will be assembled cross view to update CCF. In other words, CCF is online updated to enhance its robustness against view and appearance change. 

{\bf Third}, for a view with low confident initial tracking results ({\it e.g.}, view 3 in Figure~\ref{fig:overall}), TPN will be used to estimate its tracking result by implicitly taking into account the geometric relations among different views/cameras.

\begin{algorithm}[!t]
    \caption{Generic Multiview Tracking\label{al1}}
    \SetAlgoNoLine 
    \SetKwInOut{Input}{\textbf{Input}}\SetKwInOut{Output}{\textbf{Output}} 
 
    \Input{$\mathcal{I}_t$: input images at time $t$\;\\
        $\mathcal{G}_{t_0,t-1}$: previous trajectories of all cameras\;\\
        $f$: dynamically updated collaborative filters\;\\
        $\mathcal{B}_{t-1}$: tracking results (boxes) in last frame\;\\
        $\mathcal{Z}^c$: training samples for each view $c$\;\\}
    \Output{$\mathcal{B}_t$: tracking result for time $t$\;\\
        $\mathcal{G}_{t_0,t},f,\mathcal{Z}^c$: updated results\;\\
        }
    \BlankLine
 
    \For {~each camera $c$}{
        \For {~each scale $k$}{
            $U_k^c$ = CropImagePatch($I_t^c$, ${\bf b}_{t-1}$, $k$)\;
            $X_k^c$ = FeatureExtraction($U_k^c$)\;
            $Y_k^c$ = Correlation operation between $X_k^c$ and $f$\;
        }
        $k'= \max_{k}\{Y_{k}^c\}$\;
        Localize object ${\bf b}_t^c$ and update ${\bf g}^c$ according to $Y_{k'}^c$\;
        $q^c = Y_{k'}^c$\;
        
        \If {~($q^c\ge\tau) ~\mathrm{and}~ (t\mod 7=0)$} 
        {
           $\mathcal{Z}^c = \mathcal{Z}^c \cup X_{k'}^c$\;
        }
    }
    \If {~$t \mod 7=0$}
    {
        Update $f$ by training on all $\mathcal{Z}_c: c=1,\dots,n_c$\;
    } 
    \For {~each camera $c$}{
        \If {~$q^c < \tau$}
        {
            Update ${\bf g}^c$ using TPN (\S~\ref{TPSec})\;
            Update ${\bf b}_t^c$ accordingly\;
        }
    }
\end{algorithm}
The overall framework is summarized in algorithm \ref{al1}. More details are described in the following subsections.

\subsection{Spatial-aware Feature Extraction Network} \label{featureExtactSec}



\begin{figure}[t]
\begin{center}
   \includegraphics[width=0.96\linewidth]
   {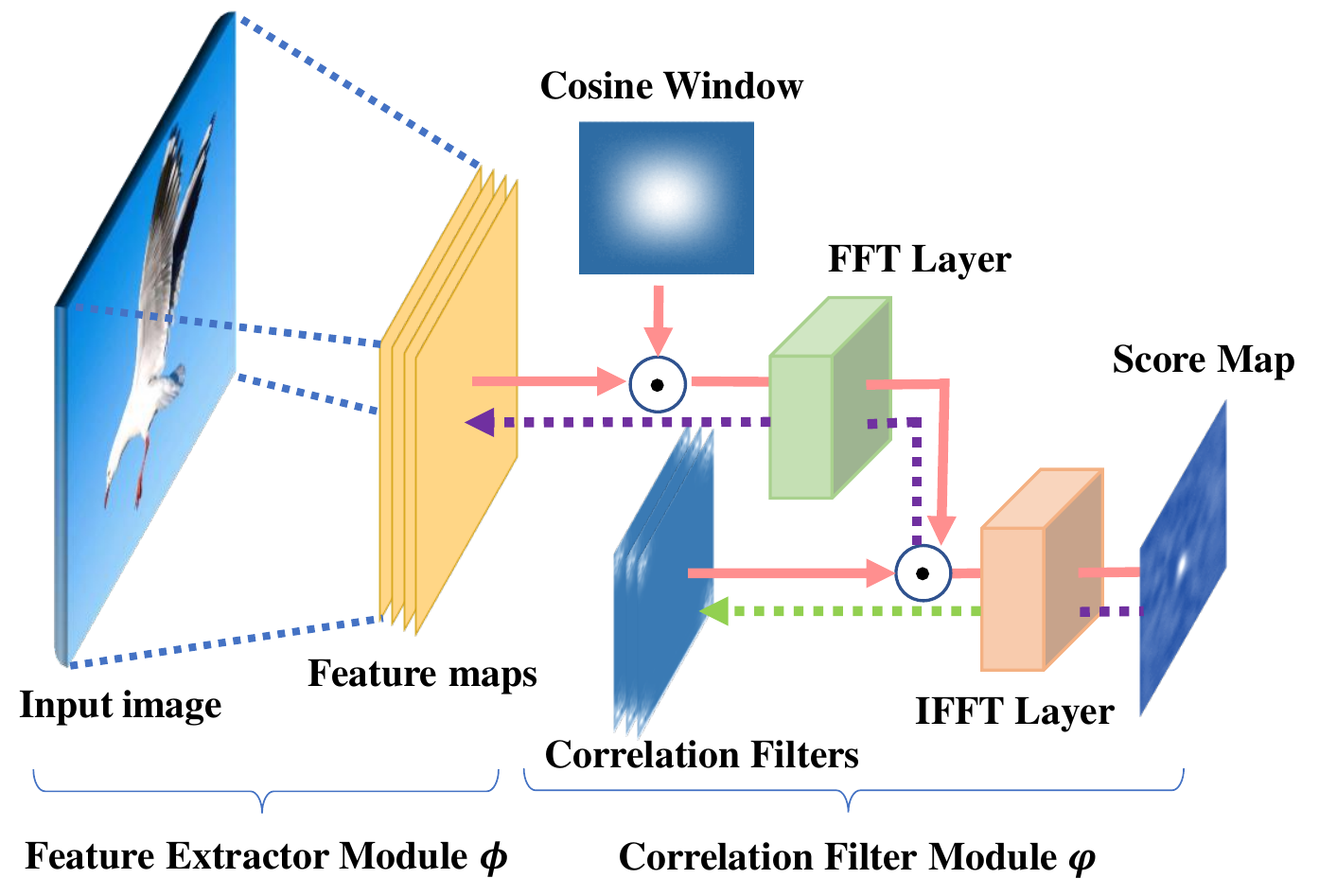}
\end{center}
   \caption{Two-stage training of feature extractors (offline). The first stage (green dotted line) is for training the correlation filters while keeping the feature extraction module fixed; while the second stage (purple dotted line) is for training the feature extraction module while keeping the correlation filters fixed.
   }
\label{fig:fenet}
\end{figure}

In order to improve the feature adaptability to correlation filter-based multiview tracking, we fine tune feature extraction model based on ResNet-50 that is pretrained on ImageNet. The offline fine tuning involves two components: the feature extractor module $\phi(\cdot)$ and the correlation filter module $\varphi(\cdot)$, as shown in Figure \ref{fig:fenet}.



We construct the training set from existing single view tracking datasets including VOT2017~\cite{Kristan2017The}, OTB100~\cite{Wu2015Object} and LaSOT~\cite{FanLYCDYBXLL19cvpr}. Different frames in a sequence can be regarded as the multi-view appearances of the same object. Data augmentation is conducted by slightly disturbing object locations. 
We randomly choose 16 sample pairs from the same sequence as a training batch. 

The training for each batch has two stages. At the first stage, the 10 training pairs are used to train the correlation filter $f_\mathrm{train}$ by minimizing the following objective function:
\begin{equation}
\begin{split}
    f_\mathrm{train}^* &= \mathop{\arg\min}_{f} \| Y^* - Y \|_2^2 + \eta\cdot\|f_\mathrm{train}\|_2^2  \\ 
    Y &= \varphi(\psi_{\mathrm{cos}}(\phi(\mathrm{X},\theta_f)),f_\mathrm{train})
\end{split}
\end{equation}
where $Y^*$ is target score map; $\theta_f$ is the parameters for feature extractor; $\eta$ is a penalty parameter; and $\psi_{\mathrm{cos}}(\cdot)$ is the Hann window function. 

After the first training stage, we obtain the optimal filter $f_\mathrm{train}^*$ of this current object. Then, we use the rest of the sample pairs to train $\phi(\cdot)$. The output response map may have some blurred noise which is supposed to be zeros when only L2 loss is used. So we add a gradient-like term to the objective function to alleviate this phenomenon. The objective function is:\\
\begin{equation}
\begin{split}
    \theta^* &= \mathop{\arg\min}_{\theta} \| Y^* - Y \|_2^2 + \lambda \|\nabla_s(Y^*) - \nabla_s(Y)\|_2^2  \\
    Y & = \varphi(\psi_{\mathrm{cos}}(\phi(\mathrm{X},\theta)),f_\mathrm{train}^*)
\end{split}
\end{equation}
where $\nabla_s(\cdot)$ is the Sobel operator. We use Adam optimizer to train our spatial-aware feature extraction networks. Note that $f_\mathrm{train}$ is only used during offline training process.

\subsection{Collaborative Correlation Filter} \label{ccfSec}

View change is a notorious issue that troubles single view trackers. Fortunately, in the multiview tracking setup, images captured from different cameras naturally provide cross-view information for building reliable tracking models. Therefore, we extend the traditional correlation filter to a {\it collaborative} one. Specifically, during tracking, we update the correlation filters online with information collaboratively collected from all sufficiently reliable views. 

Denote the training samples dynamically collected from view $c$ by $\mathcal{Z}^c=\{X_j^c\}_{j=1}^{m_c}$ with $m_c$ samples. We train a shared multiview collaborative correlation filter $f$ using all samples from different sample sets (\ie $\{\mathcal{Z}^c\}_{c=1}^{n_c}$) by minimizing the following function
\begin{equation}
E({f}) = \sum\limits_{c=1}^{n_c}\sum\limits_{j=1}^{m_c}\alpha_j^c\|X^{c}_{j} * f -Y^{c}_{j}\|^2 +\|f\|^2\quad
\end{equation}
where $Y^{c}_{j}$ denotes the score map of the $j$-th sample in the $c$-th camera, and $*$ is the convolution operation. The weights $\alpha_j^c\geq 0$ represent the importance of the $j$-th training sample of camera $c$, which is positively correlated with $q^c$ during tracking. In this formulation, training samples from all camera views contribute to filter updating, and thus enhance the robustness of the learned filters against view change.

\subsection{Trajectory Prediction Network}\label{TPSec}

Our key novelty in multiview tracking is the proposed \emph{Trajectory Prediction Network} (TPN) for handling tracking failure using cross-view trajectory prediction. Intuitively, when the target is occluded (or damaged similarly) in a \emph{target view} $b$, we can usually still reliably track the target in a different view, say, a \emph{source view} $a$. Then, based on the geometric relation between the two views, we shall be able to locate the occluded object in view $b$ from the trajectory in view $a$. The job, despite being nontrivial due to non-linearity and camera movement, is done by TPN.

\vspace{1mm}\noindent{\bf Network design.}
Denote the trajectories at time $t$ for view $a$ and view $b$ by ${\bf g}_t^a$ and ${\bf g}_t^b$, respectively. It is natural for us to find the direct mapping and prediction between them. This idea does not work in practice due to the large range of absolute coordinate of object locations. Instead, we decompose a trajectory as a sequence of between-frame movements, denoted by ${\bf r}_t^c = {\bf g}_t^c - {\bf g}_{t-1}^c $ as the the motion vector for camera $c$ at time $t$. Then, TPN aims to map from ${\bf r}_t^a$ to ${\bf r}_t^b$ at time $t$.


At time $t$, based on 3D geometrical constrains, the object position ${\bf g}^b_{t}$ in view $b$ can be transformed from its location ${\bf{g}}^a_{t}$ in $a$. Let $d_t^a$ (or $d_t^b$) and ${{\bf T}_t^a}$ (or ${{\bf T}_t^b}$) denote respectively the depth and transformation matrix for view $a$ (or $b$). We can have the following derivation
\begin{equation}
\vspace{-6mm}    
    \begin{split}
    \lambda\begin{bmatrix}
    {\bf g}_{t}^b\\
    1
    \end{bmatrix}
    &= {\bf T}_{t}^bd_{t}^a({\bf T}_{t}^a)^{-1}\begin{bmatrix}
    {\bf g}_{t}^a\\
    1
    \end{bmatrix}\\
    \lambda\hspace{-1mm}\begin{bmatrix}
    {\bf g}_{t_0}^b + \sum_i^{t} \hspace{-1mm}{\bf r}_{i}^b\\
    1
    \end{bmatrix}
    &= {\bf Q}_{t}\hspace{-1mm}\begin{bmatrix}
    {\bf g}_{t_0}^a + \sum_{i=t_0}^{t} {\bf r}_{i}^a\\
    1
    \end{bmatrix}\\ 
    \lambda\begin{bmatrix}
    {\bf r}_{t}^b\\
    1
    \end{bmatrix}
    &= {\bf Q}_{t}\hspace{-1mm}\begin{bmatrix}
    {\bf g}_{t-1}^a + {\bf r}_{t}^a\\
    1
    \end{bmatrix} \hspace{-1mm}-\hspace{-1mm} \lambda\hspace{-1mm}\begin{bmatrix}
    {\bf g}_{t_0}^b + \sum_{i=t_0}^{t} {\bf r}_{i}^b \\
    1
    \end{bmatrix} 
    \end{split}\nonumber
\end{equation}
where ${\bf Q}_{t}:={\bf T}_{t}^bd_{t}^a({\bf T}_{t}^a)^{-1}$ and $\lambda$ is for normalization, and $t_0$ is the begin time of a trajectory. Such relation between ${\bf r}_{t}^b$ and ${\bf r}_{t}^a$ motivates us to design the following  {\it Recurrent Neural Network} (RNN)-based TPN model and introduce corresponding hidden parameters in the model:   
\begin{equation}
  {\bf r}_{t}^b = \Theta_{\mathrm{pos}}( \Theta_{\mathrm{rnn}}( \Theta_{\mathrm{enc}}({\bf r}_{t}^a), {\bf p}_t),{\bf h}_{\mathrm{\mathrm{p}}} )  
\end{equation}
In the model, $\Theta_{\mathrm{enc}}(\cdot)$ is an encoder network to translate/convert the input;  $\Theta_{\mathrm{rnn}}(\cdot,\cdot)$ indicates stacked RNNs to simulate the non-linear transformation decided by ${\bf Q}_{t}$ and accumulate temporal information (\eg $\sum_i^{t} {\bf r}_{i}^b$, ${\bf g}_{t}^a$ and object movement); ${\bf p}_t$ denotes hidden states of RNN at time $t$ and it initially encodes camera matrices ${{\bf T}_t^c}$ and the initial position ${\bf g}_{t_0}^a$; ${\bf h}_{\mathrm{\mathrm{p}}}$ encodes the initial position ${\bf g}_{t_0}^b$; and $\Theta_\mathrm{pos}(\cdot)$ decodes all features to output the results.  



\begin{figure}[t]
\begin{center}
   \includegraphics[width=1\linewidth]{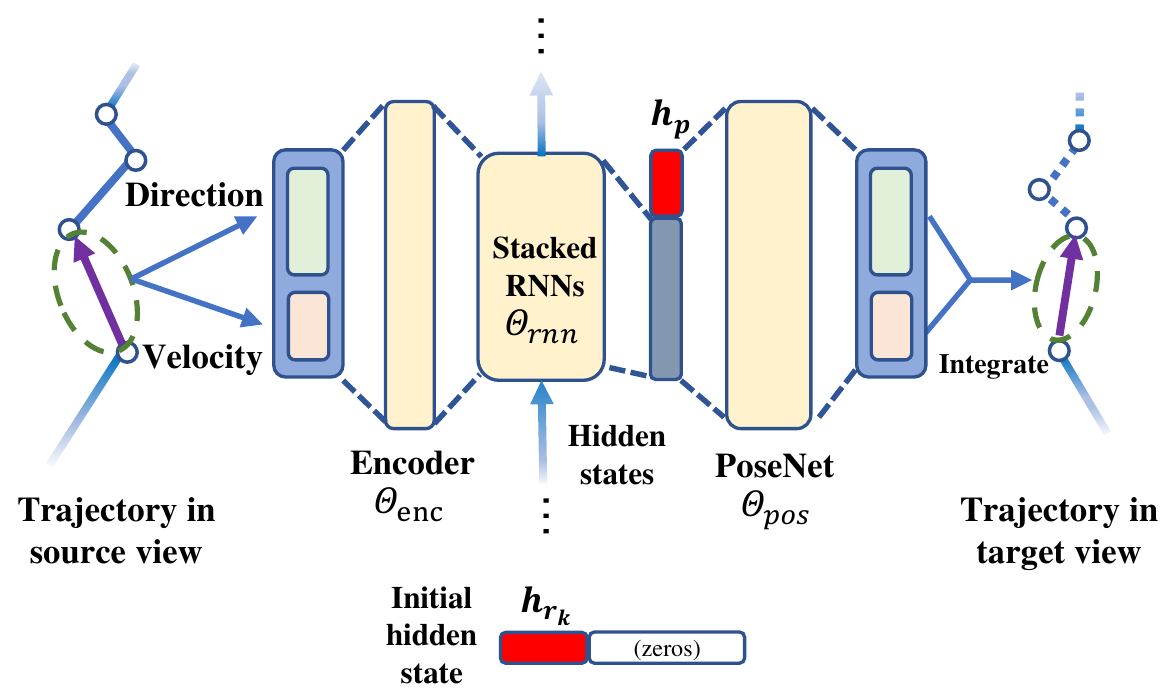}
\end{center}
   \caption{Prediction by TPN. The trajectory in the source view $a$ is decomposed into direction and velocity and arranged in a motion vector. Then the encoder $\Theta_{\mathrm{enc}}$ maps the motion vector into a 128-dimensional representation, which then passes through $\Theta_\mathrm{rnn}$ containing with 2 stacked RNN layers and 2 learnable hidden states ${\bf h}_{r_1},{\bf h}_{r_2}$. Following that, another hidden vector ${\bf h}_\mathrm{p}$ is concatenated with output of RNNs and sent to PoseNet $\Theta_{\mathrm{pos}}$ that captures the between-view geometric constrain. During prediction, only ${\bf h}_{r_1},{\bf h}_{r_2}$ and ${\bf h}_\mathrm{p}$ need to be updated. The trajectory in the occluded target view $b$ is corrected by integrating the predicted motion vector  $\mathcal{R}_{t_0,t}^b$. Here, we only illustrate a data flow at a specified time $t$ during prediction.}
\label{fig:tracjnet}
\end{figure}

The structure of TPN is shown in Figure~\ref{fig:tracjnet}. The initial hidden state ${\bf p}_{t_0}^k$ of $k$-th RNN layer consists of zero vector and a learnable hidden parameter vector ${\bf h}_{r_k}$, \ie, ${\bf p}_{t_0}^k=[{\bf h}_{r_k},{\bf 0}]$. These initial hidden states form ${\bf p}_{t_0}=\{{\bf p}_{t_0}^1,{\bf p}_{t_0}^2\}$ together, where the hidden states ${\bf p}_t$ of RNN in general encode temporal information. Moreover, we use ${\bf h}_{\mathrm{rnn}}=[{\bf h}_{r_1},{\bf h}_{r_2}]$ to represent all hidden parameters of RNN. PoseNet is a deep fully-connected network whose input includes both the output of RNN and a hidden parameter vector ${\bf h}_\mathrm{p}$. Therefore, TPN can be viewed as a decoder that parses hidden parameters to a mapping function that maps motion vectors from source view into the target view. Note that ${\bf Q}_{t}$ is treated as a dynamic transformation, and thus allows camera movement. 

In practice, estimated trajectories from a source view often contain noise that may cause unstable trajectory prediction for the target view. For this reason, we smooth the source trajectories before sending them to TPN. Specifically, the smoothed motion vector ${\bf r}_t^c$ (we abuse the notation ${\bf r}_t^c$ for conciseness) is estimated by
\begin{equation}\label{speedEq}
{\bf r}_t^c = \frac{1}{3}\hspace{-.5mm}\sum\limits_{j=0}^2\|{\bf g}_{t-j}^c-{\bf g}_{t-j-1}^c\|\ \hspace{-.85mm}\cdot\ \hspace{-.85mm} \frac{1}{3}\hspace{-.5mm}\sum\limits_{j=0}^2\frac{{\bf g}_{t-j}^c-{\bf g}_{t-j-1}^c}{\|{\bf g}_{t-j}^c-{\bf g}_{t-j-1}^c\|}.
\end{equation}
In this way, $\bf{r}_t^c$ consists of two parts, the velocity (left) and the direction (right). 

\vspace{1mm}\noindent{\bf Trajectory Dataset.}
To train and test TPN, we first prepare a trajectory dataset by collecting trajectory pairs from different kinds of camera settings and object motions. In total, 25 scenarios are used for training and 8 for testing. The data of each scenario is captured by two cameras with different relative pose constrains. The between-camera relative pose may change slightly during capturing. An object is placed in front of the cameras. We move the object or the cameras randomly in the free space so that trajectory pairs are formed in different views without occlusion. We have 30,000 frames altogether and each sequence has at least 900 frames.

\vspace{1mm}\noindent{\bf Training TPN.}
As for training, we sample $n_b$ ($n_b=100$) trajectory pairs from the 25 scenarios per batch. The $i$-th trajectory pair $(\mathcal{G}_{t_0,t_2}^{a,i},\mathcal{G}_{t_0,t_2}^{b,i})$ is chosen from 90 continuous frames, \ie, $t_2-t_0=89$ with $t_0$ randomly chosen. Using Eq.\ref{speedEq}, we get a motion vector set pair $(\mathcal{R}_{t_0,t_2}^{a,i},\mathcal{R}_{t_0,t_2}^{b,i})$, where $\mathcal{R}_{t_0,t_2}^{a,i}=\{{\bf r}_t^{a,i}\}_{t=t_0}^{t_2} $. Let ${\bf h}^i=({\bf h}^i_{\mathrm{rnn}},{\bf h}^i_{\mathrm{\mathrm{p}}})$ be the learn-able hidden parameters of $i$-th sample pair in networks and $\bf \theta$ be other parameters of networks. Our objective is to find an optimal ${\bf \theta}^*$ that:
\begin{equation}
\begin{split}
{\bf \theta}^* = \mathop{\arg\min}\limits_{\theta}\sum\limits_{i=1}^{n_t}\|\Psi(\mathcal{R}_{t_0,t_2}^{a,i},{\bf \theta},{\bf h}^i)-\mathcal{R}_{t_0,t_2}^{b,i}\|^2 \\ \quad \quad  + \lambda_2\| { \Theta}_{\mathrm{int}}(\Psi(\mathcal{R}_{t_0,t_2}^{a,i},{\bf \theta},{\bf h}^i), {\bf g}_{t_0}^{b,i}) - \mathcal{G}_{t_0,t_2}^{b,i}\|^2
\end{split}
\end{equation}
where $n_t$ is the number of training pairs; $\Psi(\cdot,\cdot,\cdot)$ denotes TPN, which takes a set of motion vector $\mathcal{R}^a$, network parameters ${\bf \theta}$ and learn-able hidden parameters ${\bf h}$ as inputs, and outputs $\mathcal{R}^b$ for the target view. $\Theta_{\mathrm{int}}(\cdot,\cdot)$ integrates motion vectors into 2D absolute positions according to given initial point. We can recover a predicted trajectory by
\vspace{-2mm}
\begin{equation}
\Theta_{\mathrm{int}}(\mathcal{R}_{t_0,t_2}^{c},~{\bf g}_{t_0}^{c}) = \Big\{ {\bf g}_{t}^c | {\bf g}_{t}^c = {\bf g}_{t_0}^{c} + \sum_{t=t_0+1}^t {\bf r}_t^c\Big\}_{t=t_0}^{t_2}
\vspace{-2mm}
\end{equation}

Since ${\bf h}^i$ is also unknown in the beginning of each batch, we divide the training process into two stages: \\
{\bf{Stage 1}}: We randomly initialize ${\bf h}^i$ and conduct network training which only optimizes ${\bf h}^i$ and fixes current $\bf \theta$ for each training batch:
\begin{equation}\label{trainhidden}
 \hspace{-1mm} {\bf h}^{i^*} \hspace{-2mm}= \mathop{\arg\min}\limits_{{\bf h}^i}
 \|\Psi(\mathcal{R}_{t_0,t_1}^{a,i},{\bf \theta},{\bf h}^i)-\mathcal{R}_{t_0,t_1}^{b,i}\|^2 + \lambda_1\|{\bf h}^i\|^2
\vspace{-2mm}
\end{equation}
where ($\mathcal{R}_{t_0,t_1}^{a,i}$,$\mathcal{R}_{t_0,t_1}^{b,i}$) is first 40 frames of ($\mathcal{R}_{t_0,t_2}^{a,i}$,$\mathcal{R}_{t_0,t_2}^{b,i}$), which means $t_1-t_0=39$.
\\
{\bf{Stage 2}}: We use ${\bf h}^{i^*}$ as initial parameters, and train networks parameter $\bf \theta$ by using training samples in a batch. 
\vspace{-2mm}
\begin{equation}
\vspace{-4mm}\begin{split}
 \theta^*,{\mathcal{H}}^{**} = \mathop{\arg\min}\limits_{{\bf \theta},{{\bf h}^i}^*}\sum\limits_{i=1}^{n_b}\|\Psi(\mathcal{R}_{t_0,t_2}^{a,i},\theta,{{\bf h}^i}^*)-\mathcal{R}_{t_0,t_2}^{b,i}\|^2 \\ + \lambda_2\|  \Theta_{\mathrm{int}}(\Psi(\mathcal{R}_{t_0,t_2}^{a,i},{ \bf \theta},{{\bf h}^i}^*), {\bf g}_{t_0}^{b,i}) - \mathcal{G}_{t_0,t_2}^{b,i}\|^2 + \lambda_1\|{\bf h}^{i*}\|^2
\end{split}\nonumber
\end{equation}
where ${\mathcal{H}}^{**}=\{{\bf h}^{i**}\}_{i=1}^{n_b}$, ${\bf h}^{i**}$ is optimized parameter for ${\bf h}^{i*}$.


We use the Rprop algorithm to optimize the network parameters. After 20 epochs of training on the trajectory dataset, we obtain $\bf \theta^*$ and finish training TPN. 

\vspace{1mm}\noindent{\bf TPN in Generic Multiview Tracking.}
During generic multiview tracking, the situation may be complicated. There may be more than one unreliable and reliable views. At time $t$, for each unreliable view $b$ (\ie ${q^b}<\tau$), we use TPN to estimate its trajectory. We also take the result of correlation filter, ${\bf g}_t^b$, into account. The corrected object's location ${\bf g}_t^{b'}$ for camera $b$ is given by:
\begin{equation} \label{equ_trajection}
\begin{split}
 {\bf g}_t^{b'} = q^\frac{b}{2} {\bf g}_t^b + \frac{1-q^\frac{b}{2}}{w} \hspace{-2mm}\sum\limits_{c, q^c\geq\tau}\hspace{-1mm} q^c \Theta_\mathrm{TP}(\mathcal{G}_{t_0,t}^c,\mathcal{G}_{t0,t1}^b)
\end{split}
\end{equation}
where $w = \sum_{c, q^c\geq\tau} q^c$ is a normalized coefficient and $\Theta_\mathrm{TP}(\cdot, \cdot)$ is a trajectory prediction function, which is TPN embedded and predicts the object location ${\bf g}_t^b$ in camera $b$.

The behavior of $\Theta_\mathrm{TP}(\cdot, \cdot)$ is defined in the following. For the input ($\mathcal{G}_{t_0,t}^c$,$\mathcal{G}_{t0,t1}^b$), $\mathcal{G}_{t_0,t_1}^c$ and $\mathcal{G}_{t0,t1}^b$ are used to train hidden parameters ${\bf h^*}$ according to Eq.~\ref{trainhidden}; After that, we can obtain  $\mathcal{G}_{t_0,t}^b=\Theta_{\mathrm{int}}(\Psi(\mathcal{R}_{t_0,t}^{c},{ \bf \theta},{{\bf h}^*}), {\bf g}_{t_0}^{b})$. Finally, we take ${\bf g}_t^b$ as the output result of $\Theta_\mathrm{TP}(\cdot, \cdot)$. $t_1$ is the last time when view $b$ is reliable. We choose 40 frames to train ${\bf h^*}$, which means $t_0 = t_1-39$. This equation builds connections among multiple cameras and guides the trajectory correction when occlusion occurs.

In reality, there can be no reliable views. In that case, we keep the last momentum of the target object in each view. Let ${\bf r}_{t_1}^c$ be the motion vector at the last reliable time $t_1$ in view $c$. ${\bf g}_{t}^{c} = {\bf g}_{t-1}^{c} + {\bf r}_{t_1}^c$.

\section{Experiments}
To evaluate methods in this task, we build a multiview tracking dataset and compare our tracker with others on it. Afterwards, we analyze the performance of our {\it Trajectory Prediction Network}. The experiment shows that the proposed networks can find out the relationship between two trajectories effectively and improve tracking performance. We also evaluate our approaches on PETS2009~\cite{ellis2009pets2009}.    

\begin{figure*}[!t]
\begin{center}
   \includegraphics[width=1\linewidth]{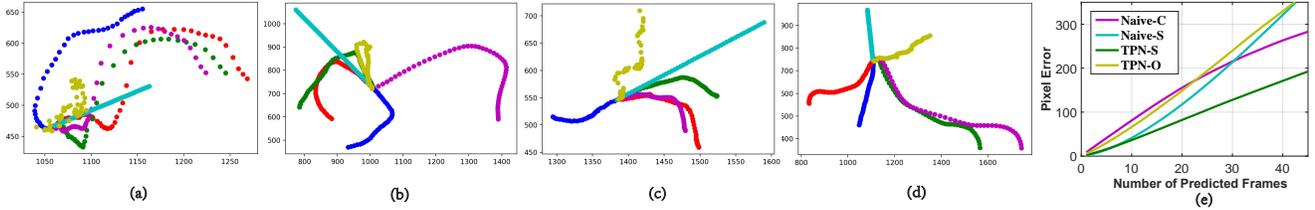}
\end{center}
   \caption{Trajectory prediction evaluation. In (a-d), blue trajectories are $\mathcal{G}_{t_0,t_1}^b$ and red trajectories are $\mathcal{G}_{t_1+1,t_2}^{b}$. We compare predictions of different methods shown in different colors. {\bf (a)} TPN-S and naive-C predict well while others deviating from the real trajectory.~{\bf (b)} The moving directions in two views are opposite, making naive-C tend to failed. {\bf (c)} TPN-S may have deviations from $\mathcal{G}_{t_1+1,t_2}^{b}$. {\bf (d)} Methods may fail when predicting at the turning point of trajectory. {\bf (e)} The plot of average prediction errors with the number of predicted frames. }
\label{fig:ex_tpn}
\end{figure*}

\subsection{Multiview Tracking Datasets}
Due to the difficulty in collecting and annotating multiview scenes, there is a serious lack of multiview tracking datasets. The PETS2009 dataset~\cite{ellis2009pets2009}, which contains sequences taking from eight cameras, is such a dataset. PETS2009, however, by itself is insufficient for convincing experimental evaluation with its low frame rate and resolution. For this purpose, we capture and manually annotate the {\it Generic Multiview Tracking Dataset} (GMTD) to facilitate relevant research and evaluation. 

GMTD contains a total of 17,571 frames and consists of 10 multiview sequences with each of them captured by two or three synchronized uncalibrated cameras, under 1080p resolution and 30fps. During data capturing, cameras are either tripod mounted or hand held. In particular, for hand-held cameras scenario, cameras may undergo small translation and rotation. Cameras are placed with different relative angles, for example, facing opposite or same directions, to form diverse trajectories in each view.

GMTD takes into account the diversity of scenarios and targets. Several different targets, including rigid ones (\eg cans, lantern and basketball) and deformable ones (\eg leaves, human and cat), were captured in the indoor, outdoor, artificial or natural scenes. During the acquisition process, a target may move under multiple camera views with more than 75\% overlap. These 10 sequences mainly cover six aspects of challenges in visual tracking, including scale variation, motion blur, deformation, background clusters, fast motion and occlusion.

The selected target is manually annotated in each sequence by axis-aligned rectangle bounding boxes. The annotation guarantees that a target occupies more than 60\% area of the bounding box. For further analysis, we also label target state in each frame as fully-visible, partially-occluded (33.73\% per sequence on average) or fully-occluded (4.78\% per sequence on average). Object is occluded by 20\% to 80\% in partially-occluded scenario. Others are fully-visible (if below 20\%) or fully-occluded (if beyond 80\%). Bounding boxes are predicted by human when the object is occluded. Some example images are in Figure~\ref{fig:GMTD}.



\subsection{Evaluation Methodology}

We evaluate our method in two ways, with (see \cite{Kristian2016A}) and without re-initialization respectively.

\vspace{1mm}\noindent{\bf With Re-initialization.}
Based on widely-used tracking performance measurements, we choose two easily interpretable measurements to evaluate methods, which are accuracy and robustness. When evaluating with re-initialization, `Accuracy' refers to the area, in percentage, of the results overlaps with the ground truth and `Robustness' is a probability of tracker failing after $S$ frames. For traditional single view single object trackers, we apply synchronous tracking evaluations on each camera view of each scene individually .

For more details, the target bounding boxes in all views will be reset to the next nearest fully-visible frame once IOU drops to zero (the tracking result has no overlapping with ground truth bounding box) in any view. The tracker will be initialized by using new bounding boxes and frame images at the same time. Let $a^c_{i,t}$ denote the IoU in view $c$ at time $t$ in scene $i$, $\mathcal{V}_i=\{ t | \forall c, a^c_{i,t}>0  \}$ the valid set. The per-scene accuracy ${\rho_{i}}$ for scene $i$ is defined as ${\rho_i}=\frac{1}{n_c|\mathcal{V}_i|}\sum_{t\in \mathcal{V}_i}\sum_{c}^{n_c} a^c_{i,t}$. We run a tracker 5 times for each scene to obtain average accuracy ${\overline{\rho_{i}}}$. Thus, the overall average accuracy $\overline{\rho}$ is obtained by the weighted average as $\overline{\rho}=\frac{1}{\sum_i |\mathcal{V}_i|}\sum_i |\mathcal{V}_i|\cdot\overline{\rho_{i}}$.


 We visualize results in accuracy-robustness(AR) plots. In AR plots, each tracker is represented as a point in terms of its overall averaged accuracy and robustness on GMTD dataset. Comparatively speaking, The tracker performs better if it is located in the top-right part of the plot and worse if it occupies the bottom-left part.

\vspace{1mm}\noindent{\bf Without Re-initialization.}
To simulate a more realistic tracking environment, we also test relevant trackers without re-initialization. Under these circumstances, the accuracy is defined as the same as re-initialization case. We denote $\delta_i = \frac{|\mathcal{V}_i|}{l_i}$ as the success rate of scene $i$ , where $l_i$ counts the total number of frames of the view in scene $i$ . Similarly, tracker is visualized in a plot with respect to average accuracy and average success rate of each scene.

We compare our method with ECO tracker~\cite{danelljan2017eco}, which is a typical example of correlation based tracker. In our experiments, all trackers' parameters are fixed and $\tau=0.5$.


\subsection{Results on the GMTD Dataset}


\begin{figure*}[!t]
\begin{center}
   \includegraphics[width=.471\linewidth]{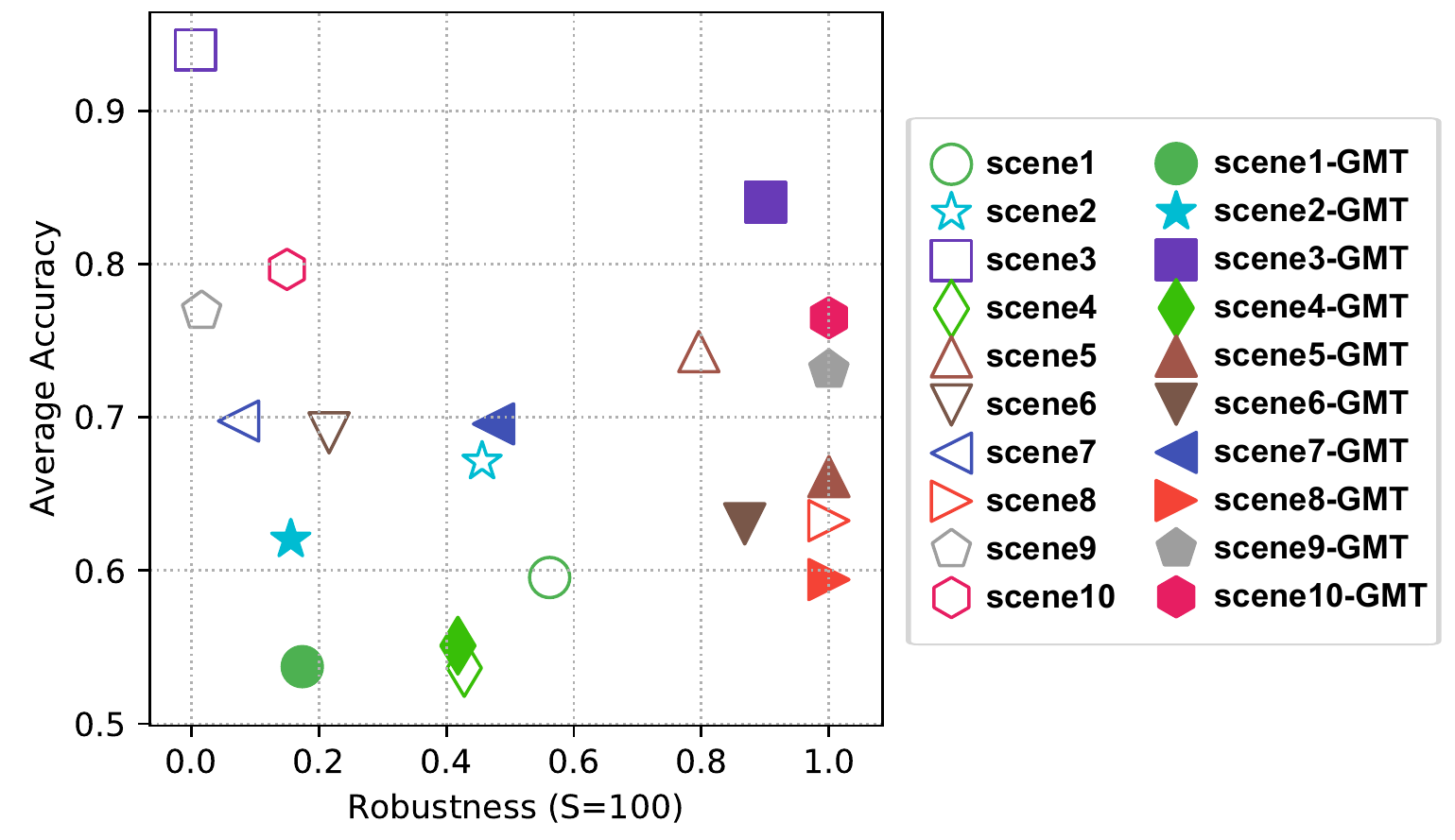}~~~
   \includegraphics[width=.471\linewidth]{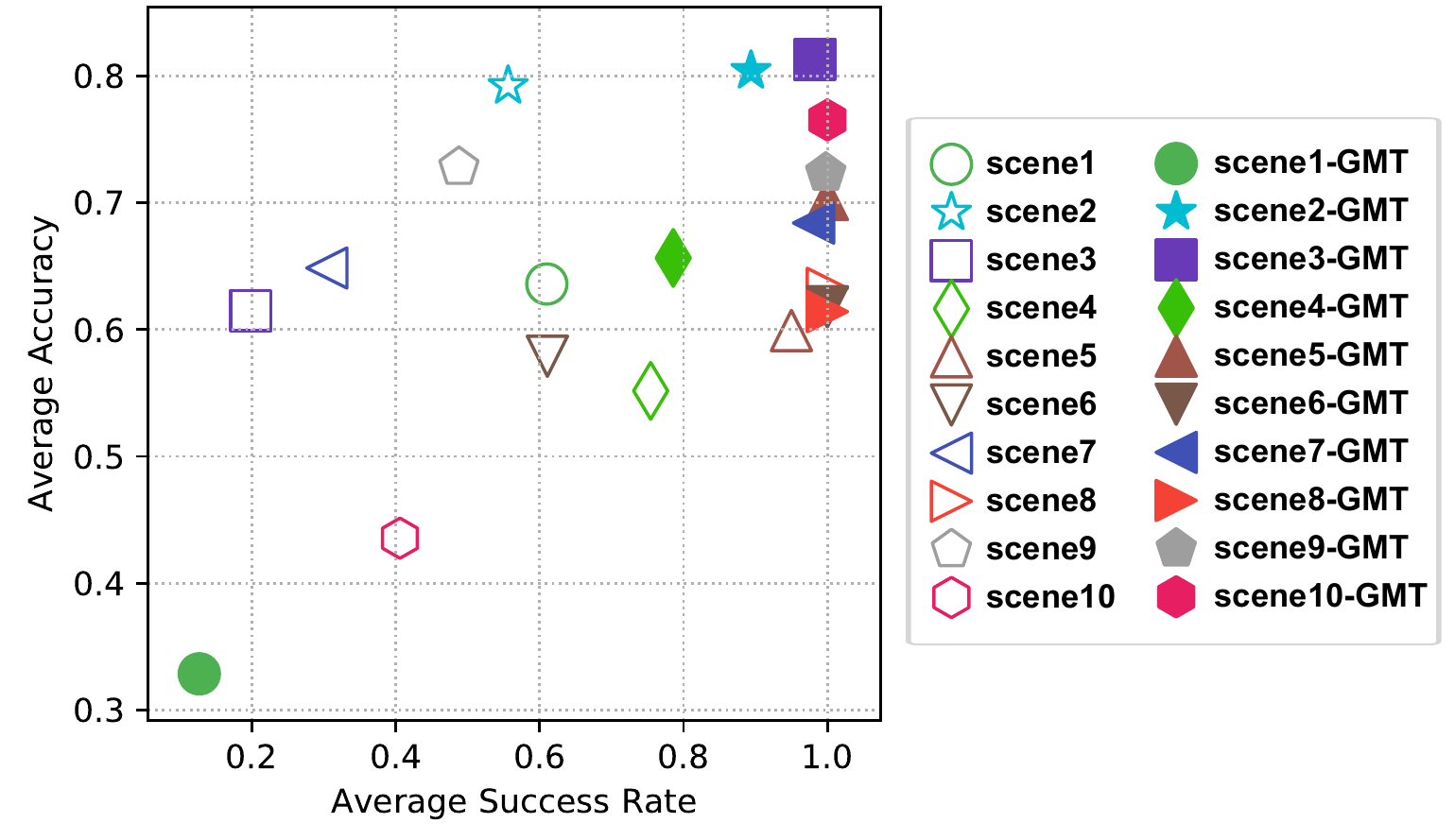}
\end{center}
   \caption{Evaluation for cases  {\bf with} re-initialization (left) and {\bf without} re-initialization (right). Hollow shapes represent results of ECO tracker, and solid ones for GMT. A tracker is better if it resides close to the top-right corner.}
\label{fig:evaluation}
\end{figure*}

\vspace{1mm}\noindent{\bf With Re-initialization.}
The overall averaged accuracy of GMT is 0.6984 and the overall average robustness is {\bf 0.7477}. On ECO tracker, the overall averaged accuracy is {\bf 0.7541} and the overall average robustness is 0.2985. We can see a huge improvement on robustness in most scenes. Results can be seen in Figure~\ref{fig:evaluation}(left).


\vspace{1mm}\noindent{\bf Without Re-initialization.}
Tracking algorithms cannot apply re-initialization due to the absence of ground truth in the real world. Thus, we also conduct evaluation without re-initialization to evaluate the long-term performance of algorithms. The results are shown in Figure~\ref{fig:evaluation}(right). 

\subsection{Trajectory Prediction Evaluation}

We evaluate our proposed TPN on trajectories' test dataset and compare TPN with naive methods. Moreover, we also test variants of TPN to show that the current structure (TPN-S) is comparatively optimal.

We provide two naive methods for trajectory prediction. One copies speeds of reference view and integrates them into the trajectory of current view (Naive-C). The other one simply repeats the last average speed of current view's trajectory (Naive-S). Variants of TPN contain the standard TPN (TPN-S) described in section \ref{TPSec}. Another variant's hidden parameters are removed and all network parameters are trained online during tracking (TPN-O). 

We simulate trajectory predictions on test dataset in online tracking between two views, $a$ and $b$, to evaluate these methods. During each simulation, we sample a trajectory pair ($\mathcal{G}_{t_0,t_2}^a$,$\mathcal{G}_{t_0,t_2}^b$) of two views from time $t_0$ to time $t_2$ ($t_2-t_0 = 89$). The trajectory pair is divided into two parts. The first part is ($\mathcal{G}_{t_0,t_1}^a$,$\mathcal{G}_{t_0,t_1}^b$), where $t_1-t_0 = 39$. We train these models by using this pair. Then we use the left trajectory $\mathcal{G}_{t_1+1,t_2}^a$ of reference view $a$ and trained model to predict the trajectory of the other view $b$, denoted $\mathcal{G}_{t_1+1,t_2}^{b*}$. After that, we use ground truth $\mathcal{G}_{t_1+1,t_2}^{b}$ to calculate the pixel distances of each frame between $\mathcal{G}_{t_1+1,t_2}^{b}$ and $\mathcal{G}_{t_1+1,t_2}^{b*}$ . In order to prevent potential variance of performance in evaluation, we repeat the simulation 1000 times and obtain the statistic average error of pixels for predicted position in each frame.

Results are shown in Figure~\ref{fig:ex_tpn}. This evaluation shows that our proposed TPN-S has the best performance in predicting trajectory from the reference view. Respectively, TPN-O, whose hidden layers are removed, suffers from over-fitting problem. Moreover, TPN-O updates all parameters online, which leads to more computational costs. There still are some failure cases for all methods, such as (d) in Figure~\ref{fig:ex_tpn}. It failed because $\mathcal{G}_{t_0,t_1}^b$ lacks sufficient information to infer the relationship between cameras.

\subsection{Results on PETS2009}

We also make a comparison on PETS2009 dataset. We trim five video clips from city center scene, which has sets of training sequences with different views. Video clips cover both sparse and dense crowd scenarios. We pick one pedestrian under two views of each video clip as targets and manually annotate ground truth bounding boxes for evaluation with re-initialization. Results are shown in Table~\ref{tab:PETS2009}.

\begin{table}
\small
\begin{center}
\caption{Evaluation results on PETS2009. Accuracy and robustness of trackers on each video clip. Overall weighted accuracy and robustness are also presented. Robustness is calculated under $S=50$.}\label{tab:PETS2009}
\begin{tabular}{r|ccccc}
\hline\hline
~ & $\rho_\mathrm{ECO}$ & $\sigma_\mathrm{ECO}$ & $\rho_\mathrm{GMT}$ & $\sigma_\mathrm{GMT}$ & \# frame/view\\
\hline\hline
clip1 & 0.708 & 0.600 & 0.695 & 0.819 & 320 \\ \hline
clip2 & 0.724 & 0.500 & 0.820 & 1.000 & 148 \\ \hline
clip3 & 0.573 & 0.340 & 0.600 & 0.380 &400 \\ \hline
clip4 & 0.684 & 0.250 & 0.674 & 0.133 &170 \\ \hline
clip5 & 0.681 & 0.450 & 0.630 & 0.581 &403 \\
\hline
Overall & {\bf 0.662} & 0.434 & 0.661 & {\bf 0.568} &- \\
\hline
\end{tabular}
\end{center}

\end{table}

\section{Conclusion}
In this paper we propose a novel \emph{generic multiview tracker} (GMT) for visual object tracking with multi-camera inputs. Unlike most previous multiview tracking systems, our GMT requests no prior knowledge about the tracking target, allows camera movement, and is calibration free. Our GMT has two novel components, a cross-view \emph{trajectory prediction network} and \emph{collaborative correlation filter}, which are effectively integrated with a correlation filter tracking framework. For evaluation, we contribute a \emph{generic multiview tracking dataset} to alleviate the lack of proper multiview tracking benchmarks. In our carefully designed experiments, the proposed tracking algorithm demonstrates advantages over state-of-the-art tracking algorithms.

{\small
\bibliographystyle{ieee}
\bibliography{egbib}
}

\end{document}